\documentclass[sigconf]{acmart}

\AtBeginDocument{%
 }

\copyrightyear{2024}
\acmYear{2024}
\setcopyright{rightsretained}
\acmConference[HRI '24]{Proceedings of the 2024 ACM/IEEE International Conference on Human-Robot Interaction}{March 11--14, 2024}{Boulder, CO, USA}
\acmBooktitle{Proceedings of the 2024 ACM/IEEE International Conference on Human-Robot Interaction (HRI '24), March 11--14, 2024, Boulder, CO, USA}
\acmDOI{10.1145/3610977.3634929}
\acmISBN{979-8-4007-0322-5/24/03}

\makeatletter
\gdef\@copyrightpermission{
  \begin{minipage}{0.3\columnwidth}
   \href{https://creativecommons.org/licenses/by-nc-sa/4.0/}
   {\includegraphics[width=0.90\textwidth]{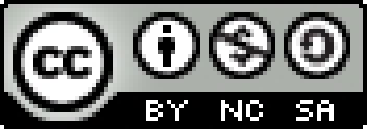}}
  \end{minipage}\hfill
  \begin{minipage}{0.7\columnwidth}
   \href{https://creativecommons.org/licenses/by-nc-sa/4.0/}{This work is licensed under a Creative Commons Attribution-NonCommercial-ShareAlike International 4.0 License.}
  \end{minipage}
  \vspace{5pt}
}
\makeatother

\begin{document}

\title{Design~and~Evaluation~of~a~Socially~Assistive~Robot Schoolwork~Companion~for~College~Students~with~ADHD}


\author{Amy O'Connell}
\affiliation{%
  \institution{University of Southern California}
  \city{Los Angeles}
  \country{USA}}
\email{ao71627@usc.edu}

\author{Ashveen Banga}
\affiliation{%
  \institution{University of Southern California}
  \city{Los Angeles}
  \country{USA}}

\author{Jennifer Ayissi}
\affiliation{%
  \institution{University of Southern California}
  \city{Los Angeles}
  \country{USA}}

\author{Nikki Yaminrafie}
\affiliation{%
  \institution{University of Southern California}
  \city{Los Angeles}
  \country{USA}}

\author{Ellen Ko}
\affiliation{%
  \institution{University of Southern California}
  \city{Los Angeles}
  \country{USA}}

\author{Andrew Le}
\affiliation{%
  \institution{University of Southern California}
  \city{Los Angeles}
  \country{USA}}

\author{Bailey Cislowski}
\affiliation{%
  \institution{University of Southern California}
  \city{Los Angeles}
  \country{USA}}

\author{Maja Matari\'c}
\affiliation{%
  \institution{University of Southern California}
  \city{Los Angeles}
  \country{USA}}


\renewcommand{\shortauthors}{Amy O’Connell et al.}

\begin{abstract}
College students with ADHD respond positively to simple socially assistive robots (SARs) that monitor attention and provide non-verbal feedback, but studies have been done only in brief in-lab sessions. 
We present an initial design and evaluation of an in-dorm SAR study companion for college students with ADHD. This work represents the introductory stages of an ongoing user-centered, participatory design process.
In a three-week within-subjects user study, university students (N=11) with self-reported symptoms of adult ADHD had a SAR study companion in their dorm room for two weeks and a computer-based system for one week. 
Toward developing SARs for long-term, in-dorm use, we focus on 1) evaluating the usability and desire for SAR study companions by college students with ADHD, and 2) collecting participant feedback about the SAR design and functionality.
Participants responded positively to the robot; after one week of regular use, 91\% (10 of 11) chose to continue using the robot voluntarily in the second week. 
\end{abstract}





\maketitle

\section{Introduction}

\begin{figure}[t]
  \centering
  \includegraphics[width=\linewidth]{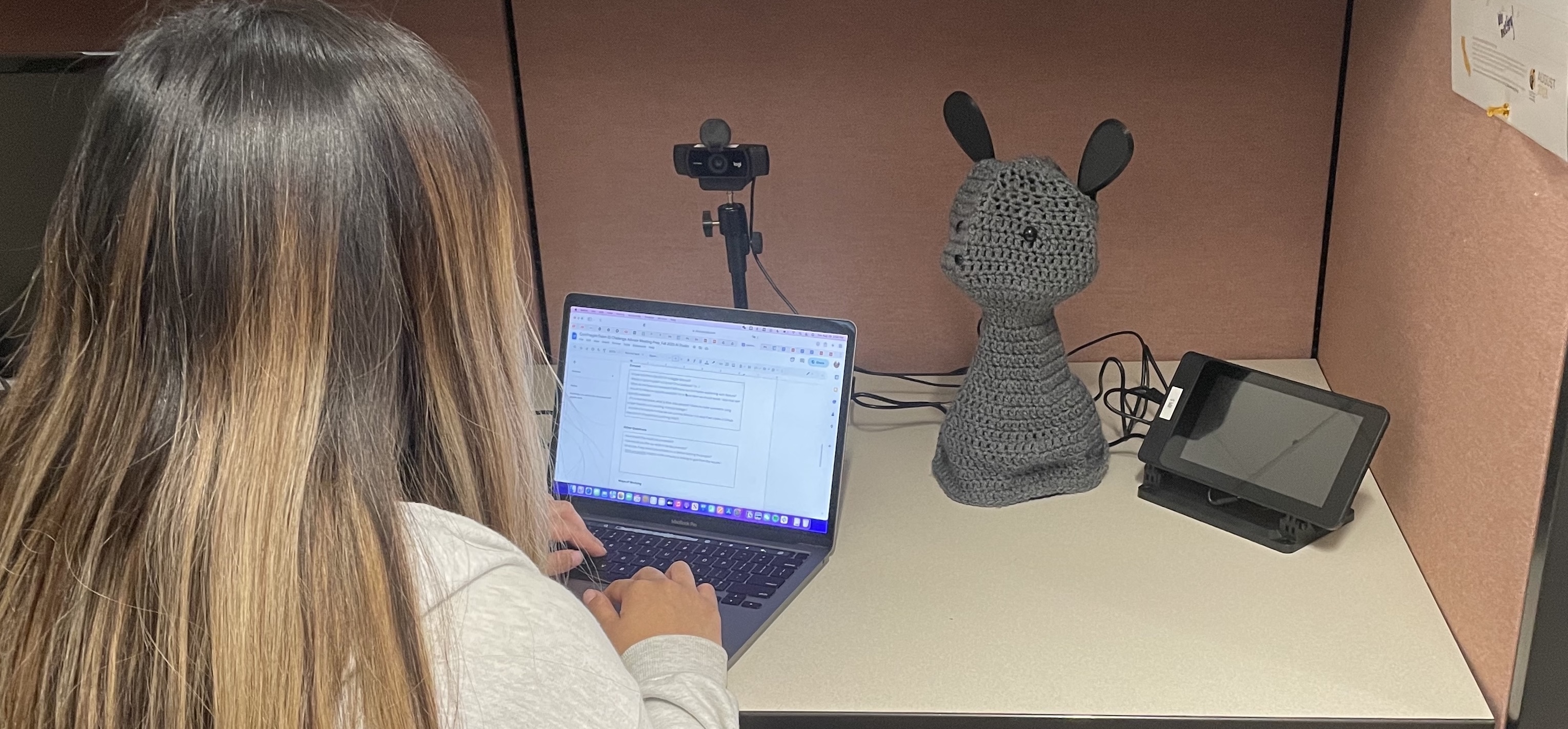}
  \caption{Blossom and the study system, including the tripod-mounted webcam and touch screen interface}
  \Description{Photograph of a student seated at a desk using a laptop. On the desk are the robot, webcam, and touch screen.}
  \label{fig:studying}
\end{figure}

College is challenging for all students, but especially for students with Attention Deficit Hyperactivity Disorder (ADHD). Students with ADHD experience more difficulty adjusting to college and lower performance than their peers without ADHD, evidenced by lower GPAs and graduation rates ~\cite{blase2009self}. 
This work seeks to develop an in-dorm SAR study companion that performs idle motions to passively support college students with ADHD during schoolwork tasks. Through a three-week, within-subjects study, we evaluated schoolwork habits in a sample (N=11) of college students with clinically significant symptoms of ADHD who had a socially assistive robot study companion placed in their dorm rooms. The three study conditions involved a fixed study time requirement (250 min) without a robot, the same amount with a robot, and optional no minimum study time with a robot.

91\% of participants used the system at least once in the third condition when not required to do so, indicating that in-dorm SAR study companions are viable tools for a subset of students with ADHD. The system had an average score of 83.86 on the System Usability Scale (SUS) ~\cite{bangor2008empirical}, and there was a strong correlation between participants' SUS scores and the amount of time they studied with the robot beyond the required amount.

\section{Related Work}

This work draws from prior research on ADHD, education, social psychology, and human-robot interaction (HRI). In this section, we introduce prior work on procrastination in ADHD and the role of co-presence and social support in facilitating action for people with and without ADHD. We connect these ideas to our work developing SAR study companions for college students with ADHD. We then explore prior technical contributions on assistive technology for students with ADHD and in-dorm robots for other populations of college students and how they relate to our work.

\subsection{Procrastination and Executive Functions in College Students with ADHD}

Studies of undergraduate students with ADHD show a positive correlation between ADHD symptoms of inattention and general procrastination ~\cite{niermann2014relation}, mediated by executive functions (EFs) ~\cite{bolden2020tomorrow}. EFs are a set of cognitive processes involved in taking self-directed actions that contribute to self-regulation; they include planning, emotional and motivational regulation, goal-directed actions, and inhibition ~\cite{barkley2012executive}. College students with ADHD self-report procrastination and related difficulties, such as delaying getting started on tasks and doing unrelated activities, as well as anxiety in response to procrastination ~\cite{gray2016symptom}.
One strategy to support students with ADHD may involve using the presence of others to accomplish tasks, referred to by the neurodivergent (ND) community as "body doubling." Despite receiving extensive media attention ~\cite{chadd2022bodydouble,cnn2023bodydouble}, body doubling has appeared in only a few peer-reviewed publications ~\cite{ogrodnik2023exploring, eagle2023proposing}. 
In a recent study, Eagle et al. ~\cite{eagle2023proposing} investigated how ND individuals define and use body doubling and proposed it as a way for assistive technology to support task completion/initiation for ND individuals. They collected survey responses from 220 participants (139 identified as having ADHD) to form a community-sourced definition of body doubling:
"\textit{having someone in the room (n = 127) or on a call/chat (n = 27) in order to accomplish a task (n = 65) or be productive (n = 38). The second person may be doing a different task (n = 65) or a similar one (n = 13), and it is a form of accountability (n = 23) and helps you stay on task (n = 21)}" ~\cite{eagle2023proposing}. The authors proposed a 2-axis representation of body doubling; one axis represents mutuality, i.e., the body double's level of awareness, ranging from performance/accountability on one end to ambient companionship on the other, and the other axis represents space and time, ranging from no defined time/place on one end to the same time and/or place shared by the individual and body double on the other.

Following the recommendations of Eagle et al., we explored how socially assistive robots can act as body doubles, providing ambient companionship in a shared time and place to support college students with ADHD in schoolwork tasks.

\subsection{Social Facilitation and Inhibition}

Improved performance resulting from the presence of other individuals relates to the theory of social facilitation. This theory, first described by Triplett ~\cite{triplett1898dynamogenic}, then Allport ~\cite{allport1924social}, suggests that people perform better on low-complexity tasks when the task is performed alongside another person but perform worse on high-complexity tasks under the same condition (referred to as social inhibition). Zajonc later proposed the drive theory of social facilitation, stating that the mere presence of another person brings about enhanced drive and elicits dominant responses (responses with the greatest habit strength) \cite{zajonc1965social}. Zajonc suggested that less effective dominant responses to complex, difficult tasks could explain why the presence of others facilitated performance on familiar and easy tasks but hindered it on novel or difficult tasks. He also suggested that the presence of an audience enhances the performance of well-practiced responses but hinders new skill acquisition.
Riether et al. ~\cite{riether2012social} conducted a study of the role of social facilitation in robots and measured participant performance on simple and complex tasks in the presence of a human, an anthropomorphic robot, and alone, and found that participants performed significantly better with a robot than alone. There was no significant difference in performance between the robot and human conditions. Wechsung et al. ~\cite{wechsung2014investigating} further found that, between non-anthropomorphic and very human-like robots, participants experienced decreased performance on a complex task in the presence of the very human-like robot compared to the non-anthropomorphic robot, supporting that social inhibition may be more prevalent in interactions with highly anthropomorphic robots. Participants performed best on both tasks with the non-anthropomorphic robot present, meaning that social facilitation was not observed. In this work, we explore how social facilitation and inhibition can be applied in a real-world interaction between a SAR study companion and college students with ADHD.

\subsection{Designing for Users with ADHD}
Our work draws on the aforementioned theories to design a socially assistive robot study companion to assist college students with ADHD by providing companionship as they work on schoolwork tasks. The methodology is informed by the recommendations of Spiel, Williams, Hornecker, and Good ~\cite{spiel2022adhd} on user-centered and participatory design for populations with ADHD, first by involving four researchers (50\%) that identify as part of the intended user population of college students with ADHD, and second by testing a low-fidelity version of the interaction among a sample of college students with ADHD symptoms to get comprehensive feedback and suggestions to inform future development of the robot. 

\subsection{Technologies for students with ADHD}

In past work, researchers have made significant strides in exploring the potential of technology, including robots, to support students and children with ADHD. For instance, Adams et al. ~\cite{adams2009distractibility} explored the use of virtual reality technology to create a virtual classroom environment with programmed distractions, shedding light on the attentional challenges that children with ADHD face during school tasks and how technology can mediate these challenges. Fewer studies have explored the applications of SARs for students with ADHD. Berrezueta-Guzman et al.  ~\cite{berrezueta2021assessment} created Atent@, a Robotic Assistant (RA), and a smart home environment that utilized data from two IoT devices (chair and desk) to support children with ADHD in their homework activities. Zuckerman et al. ~\cite{zuckerman2016kip3} designed Kip3, a social robotic device that employs a tablet-based Continuous Performance Test (CPT) to assess inattention and impulsivity in college students with ADHD. Their initial evaluation suggested that Kip3 has the potential to help students regain focus, but questions remain about its long-term effectiveness and its ability to identify inattention in more complex, real-world situations. Our work begins to address questions of long-term effectiveness by deploying a study companion SAR to the dorms of college students with ADHD for two weeks, and a study system with no robot for a control period of one week. By allowing participants to work on their schoolwork tasks with the robot, we gained further insights into participants' needs and preferences for the robot's functionality and design.

\subsection{In-Dorm Robots}
Few studies have attempted to deploy robots into the dorms of college students. Abendschein, Edwards, and Edwards ~\cite{abendschein2022novelty} gave robotic cats to 9 college students for six weeks, then conducted a qualitative analysis of interviews with participants to assess the lasting novelty of an in-dorm robot. They found that novelty and use of the robot companion decreased over the six-week period. We will evaluate if the same downward trend of robot use exists for study companion robots. Jeong et al. ~\cite{jeong2020robotic} deployed 35 Jibo robots to college students' dorm rooms to perform a daily positive psychology intervention with the participant. They found that after completing the study, participants' psychological well-being, mood, and readiness to change behavior improved significantly. Our study followed similar recruitment and deployment methods but with duration of use and perceived usefulness as the primary measures of success.

\section{Methods}

\subsection{Research Questions}

 Consistent with the exploratory nature of this study, we sought to answer the following research questions:\\
{\bf RQ1:} Will college students with ADHD find a SAR useful as a study companion, as measured by quantitative surveys and voluntary use?\\
{\bf RQ2:} What features of the system will students with ADHD like and dislike during study sessions? How would students like the robot to behave, and how common are those preferences among students?\\
{\bf RQ3:} What features could be added to make the robot study companion more useful to students with ADHD?

\subsection{SAR Study Companion System}
We aimed to involve end users in the design process early while simultaneously creating an initial design sophisticated enough to give users an idea of how the study companion might look and function. Therefore, we chose to use an existing robot embodiment that could be easily adapted for our deployment.
Specifically, we used the Blossom open-source 3D-printed robotics research platform developed by Suguitan and Hoffman \cite{suguitan2019blossom}, which is inexpensive and could be quickly fabricated at the scale needed for this deployment. 
We chose a grey crocheted exterior with button features, similar to the crocheted exterior described in \cite{suguitan2019blossom}, to give Blossom a simple and engaging appearance that is not too distracting. We used a basic version of the Blossom robot in this study to create a minimum working design for initial user testing and obtaining user feedback on the robot's embodiment to inform future iterations of the physical robot design.

\begin{figure}[ht]
  \centering
  \includegraphics[width=\linewidth]{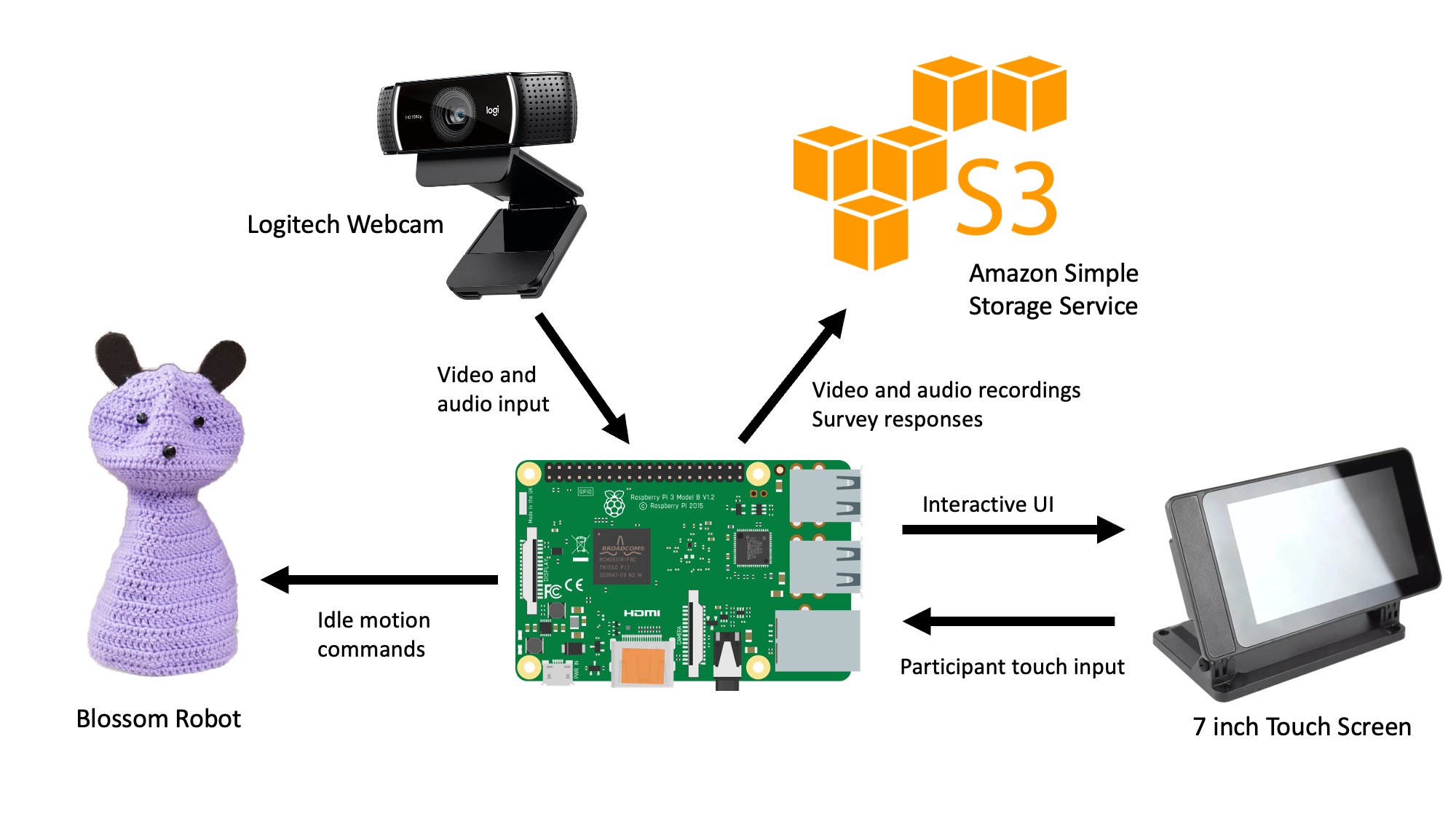}
  \caption{The Blossom robot, Logitech webcam, and 7-inch touch screen UI connected to a Raspberry Pi 4 computer}
  \Description{A diagram of the system. The touch screen sends user inputs to the Raspberry Pi. The Raspberry Pi sends idle motion commands to the robot and sends recordings and system data to AWS S3 cloud storage.}
  \label{fig:studysystem}
\end{figure}

The entire system, pictured in Figures \ref{fig:studying} and \ref{fig:studysystem}, consisted of a Blossom robot, a tabletop tripod and webcam, and a Raspberry Pi 4 computer connected to a 7-inch touch screen display. A simple user interface (UI) displayed on the touch screen allowed the participant to start, pause, continue, and end schoolwork sessions. They were also able to preview the webcam input before starting a session to confirm that they were visible in the video frame. The system recorded a log of UI button press events throughout the deployment.

\subsection{Interaction Design}

During an active schoolwork session, the touch screen displays a 25-minute timer that counts down to 0:00. When 25 minutes have elapsed, or if the user elects to end the session early, the video and audio recorded from the webcam are stored in AWS S3 cloud storage. The 25-minute duration was chosen because prior research has shown that students with ADHD commonly use the Pomodoro technique ~\cite{pomodoro}, which involves working in 25-minute sessions followed by 5-minute breaks to ensure that focus-intensive tasks are interspersed with short breaks ~\cite{kreider2019strategies, fichten2022apps}. This method is consistent with the "time on-time off" approach to allocating schoolwork time that many practitioners recommend for students with ADHD ~\cite{ofiesh2015voices}.
To give users a basic idea of how the robot might behave during a study session, we created a simple behavior policy that involved executing one of three types of hard-coded motions at random intervals throughout the study session. While a web camera recorded video and audio of the sessions for post-study analysis, the participant's actions did not influence Blossom's behavior. This was communicated to participants at their system setup appointment. During sessions, they were permitted to work on any schoolwork task, including any assignments or study activities. They were not permitted to engage in activities that were not directly related to their classes, such as leisure or extracurricular activities.  

For the duration of each study session, the robot performed a set of idle motions to maintain a lifelike and friendly presence. Studies have found that idle motions, small, lifelike movements that robots and agents perform during periods of inactivity ~\cite{cuijpers2015motions}, can help robots appear more friendly ~\cite{asselborn2017keep} and entertaining ~\cite{neggers2021self}. Following the idle motion designs described in ~\cite{cuijpers2015motions}, three types of idle motions were implemented for this study: \textit{gaze shift}, \textit{posture sway}, and \textit{sigh}. We chose to follow the idle motions described by ~\cite{cuijpers2015motions} because they were clearly defined at the actuator level, making them easy to replicate on Blossom, and had been validated as portraying low social verification for a robot in a similar task companion context. Sighs were implemented by actuating the head to its maximum height over a 2-second period then lowering the head back to a neutral height over another 2-second period. The sighing motion repeated every 60 seconds for the duration of the interaction. Idle gaze shifts were implemented by actuating the head pitch and whole body yaw to one of three predefined values (pitch: chin down, neutral, chin up, yaw: turn left, look straight ahead, turn right) over a 0.5-second period. Idle gaze shifts were executed at random intervals that varied between 15 and 22 seconds. The posture sway motions were implemented by actuating the head roll to one of three predefined positions (tilt left, tilt right, neutral) over a 1-second period. Posture sway motions were executed at random intervals that ranged between 20 and 30 seconds. All idle motions were implemented as actuations of each of Blossom's four motors to hard-coded goal positions. Two of the authors with ADHD completed test sessions with the robot to fine-tune the speed and exaggeration of the idle motions to avoid motions that would be too invasive or distracting during a study session.

\subsection{In-Dorm User Study}

To evaluate the study companion robot in a dorm environment, we completed a user study, deploying study systems to 11 college students with clinically significant ADHD symptoms.

\subsubsection{Study Design}

To evaluate the study companion robot's performance in long-term, in-dorm conditions, we conducted a three-week within-subjects user study, in which each week corresponded to one of the following three conditions:\\
\textbf{Condition A:} Participants were asked to complete a minimum of 250 minutes of schoolwork (10 full sessions) with the Raspberry Pi, touch screen interface, and webcam, but no robot. The touch screen interface was identical to conditions B and C.\\
\textbf{Condition B:} Participants were instructed to complete a minimum of 50 minutes of schoolwork (equal to 2 full sessions) per day with the robot for 5 days during the week.\\
\textbf{Condition C:} Participants were given no minimum number of sessions to complete and permitted to leave the cover over the webcam during schoolwork sessions with the robot.

To control for ordering effects among the three study conditions, the participants were separated into two groups that determined the order in which they proceeded through the study conditions. Six participants were assigned to condition A in the first week, condition B in the second week, and condition C in the third week. The other five participants were assigned to condition B in the first week, condition C in the second week, and condition A in the third week of the study. Participants began the study on different days across one week. The start and end of each week was determined based on the date that the participant began the study. Condition C always followed condition B to avoid re-introducing novelty and learning effects in week C. 
The study was structured to encourage participants to practice using the robot daily in week B, so we could examine if daily use continued voluntarily in week C. Participants were asked to complete at least 250 minutes of schoolwork with and without the robot to collect an adequate amount of session data for post-study analysis of video and audio features.

\subsubsection{Participants}
We recruited participants for the study by sending out an initial online screening survey for interested students through university mailing lists. In the survey, candidates answered a set of questions and completed the Adult ADHD Self Report Scale (ASRS) \cite{kessler2005asrs} to determine their eligibility to participate.

Participants were selected according to the following inclusion criteria: currently enrolled as a university student, 18 years of age or older, normal or corrected normal vision and hearing, proficient in English, have a private workspace in their residence where they primarily complete most of their schoolwork, and have a score of at least 4 on the ASRS. Before taking the survey, candidates read and signed a statement of consent to collect their information for screening. Candidates that met the study criteria were emailed in the order in which their responses were received. They received a brief description of the study procedure, and those who confirmed their interest were sent a link to schedule an initial setup appointment. 
Due to the narrow inclusion criteria and time-intensive study procedure, we were able to recruit only a small sample of twelve participants for the study. We opted to proceed with this small sample rather than relax the inclusion criteria and recruit members that do not represent the intended population of college students with ADHD, following Spiel et al.'s ~\cite{spiel2022adhd} recruiting recommendations for populations with ADHD. A sample size of twelve is also comparable to that of other long-term in-dorm HRI studies ~\cite{randall2019more, zhao2022let, abendschein2022novelty}. One participant (P11) dropped out in the final week of the study; eleven participants completed the study.
Participants that completed the study identified as: 6 Female, 5 Male; 7 Asian, 1 African-American,  2 Hispanic Latino, 1 did not disclose; the ages ranged from 18 to 25 (M=21, SD=2.31). Their current level of education being pursued was: 7 Bachelor's (undergraduate), 4 Master's (graduate). Participants' majors were: 2 Psychology, 1 Communication, 1 Neuroscience, 1 Cognitive Science, 1 Biochemistry, 1 International Relations, 1 Computer Science, 1 Human Biology, 1 Health Promotion and Disease Prevention, 1 Health and Human Sciences, 1 Machine Learning, 1 Electrical Engineering. 5 participants reported having previously used a time tracking productivity app to complete schoolwork. Participants received a \$135 digital Amazon gift card upon completing the study.

\subsubsection{Procedure} 

The study was approved by the University Institutional Review Board (IRB \#UP-22-01073).
Participants who met the inclusion criteria were invited to schedule a setup appointment where a researcher traveled to the student's residence and set up the system. The participants reviewed a consent form, consented to participate in the study, then completed a pre-study survey. After setting up the system, the researcher showed the participant how to use the system then explained the study procedure.

All participants were asked to complete a short post-session questionnaire after each study session completed with the system. Between conditions A and B+C, a researcher returned to the participant's dorm to either set up a robot and connect it to the study system or collect the robot, depending on the starting condition. The study setup remained unchanged in weeks B and C (study system + robot); seven days after starting condition B, participants received an email notifying them to begin week C and a link to complete a mid-study survey.

Upon completing the third week of the study, participants were given the option to have a researcher travel to their dorm to collect the system or to break down the system themselves and return it to the research lab. In both cases, during the final appointment, participants completed a post-study questionnaire and a semi-structured interview about their experience.

\subsubsection{Measures}

In the pre-study survey, participants answered questions about their demographic information, degree program and area of study, the amount of credits they were currently enrolled in, and their expected graduation year. They also completed the Executive Skills Questionnaire-Revised (ESQ-R) ~\cite{strait2020refinement}, and the Negative Attitude towards Robots Scale (NARS) ~\cite{nomura2006experimental}.
In the post-session questionnaire, participants wrote a description of the task they worked on during the session and filled out the NASA Task Load Index (TLX) ~\cite{hart2006nasa} for that task. The TLX questionnaire is a self-report scale that estimates the cognitive load based on participant ratings.
On the mid-study survey, participants repeated the NARS questionnaire.

In the post-study survey, participants completed the NARS and ESQ-R questionnaires. Additionally, the post-study survey included some background questions about the participant's prior experiences, such as their experience with productivity apps. Finally, a researcher conducted semi-structured interviews with each participant to gather more information about their experience with the robot. These interviews were conducted by one researcher who followed a script of questions, asking unscripted follow-up questions based on the participant's responses. Participants were asked what they liked and disliked about Blossom and what they thought of Blossom's behavior during the study sessions. The researcher then asked about any changes the participant would make to the robot to improve it and what features they think would be useful in an in-dorm robot. During this ideating stage, participants were encouraged not to worry about whether their ideas were practical or technologically feasible.

\subsubsection{Analysis}
We used the log of UI events from each system to extract information about the study sessions, such as the total amount of time each participant spent in active sessions with the system under each condition and when the sessions took place.

We calculated the SUS scores for each participant from the post-study survey. 
We performed paired Wilcoxon signed-rank tests on participants' pre-study and post-study ESQ-R scores. We performed a Wilcoxon signed-rank test between the NASA-TLX scores of all participants during condition A and the combined NASA-TLX scores of all participants during conditions B and C. We performed a repeated measures ANOVA test between participants' pre-, mid-, and post-experiment NARS scores.

\subsection{Analysis of Interviews}

Audio recordings of the post-study interviews were transcribed with OpenAI's Whisper speech recognition model ~\cite{radford2023robust}, then verified by one author for correctness and separated into individual sentences. To eliminate the potential of introducing bias, two coders with no prior involvement in the project separately reviewed each transcript and identified emerging themes. The coders then met to discuss their individual findings and formed a unified list of mutual themes. They reviewed the transcripts and coded each sentence according to each theme on the unified list to identify the prevalence of each theme across all participants. Codes were assigned with inter-rater reliability (mean Cohen's $\kappa$ = 0.649). Finally, researchers identified the themes related to each research question.  

\section{Results}

\subsection{Quantitative Results}

The participants gave the system an average SUS score of 83.864, earning an "A" in usability (80.3 or higher). Spearman's rank correlation was computed to assess the relationship between SUS score and total time the participant used the study system under condition C. There was a positive correlation between the two measures (r(9) = 0.693, p = 0.018), meaning perceived system usability was a reliable predictor of continued use.

\begin{figure}[ht]
  \centering
  \includegraphics[width=\linewidth]{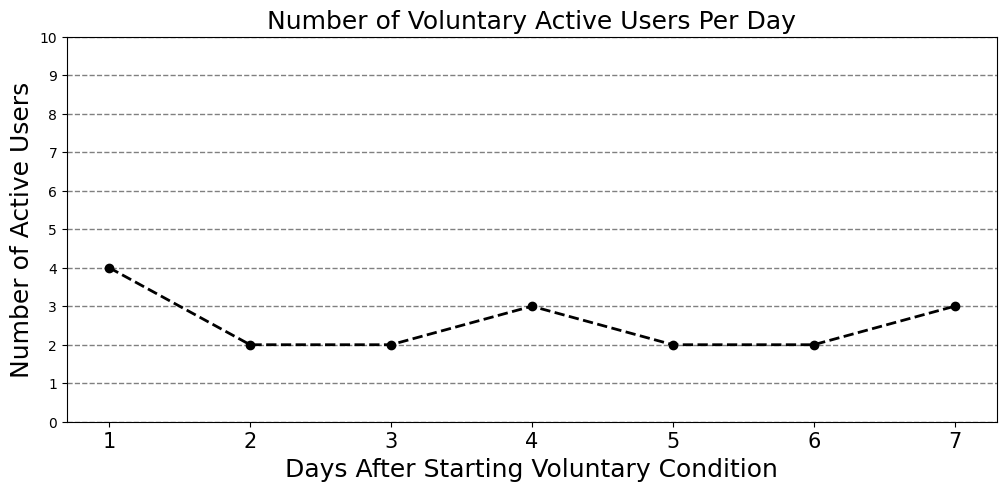}
  \caption{Daily use of the in-dorm study companion robot under condition C (voluntary use)}
  \Description{A line plot that shows the number of participants that used the robot on each day of condition C. Values are 4,2,2,3,2,2,3.}
  \label{fig:users_per_day}
\end{figure}

\begin{figure}[ht]
  \centering
  \includegraphics[width=\linewidth]{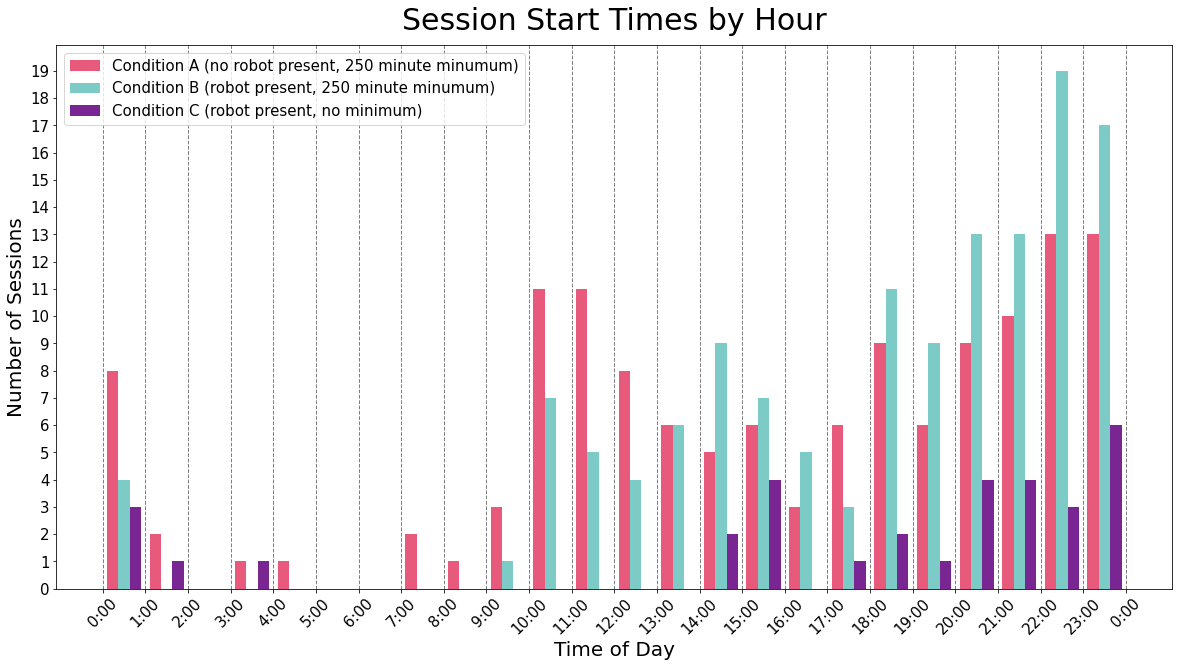}
  \caption{Session start times, by hour, in each condition}
  \Description{A histogram that shows the number of sessions initiated in each hour over a 24-hour period in each condition. Periods with the highest values are 21:00 to 00:00 and 10:00 to 11:00}
  \label{fig:sessions_per_hour}
\end{figure}

Of 11 college students who participated in the study, 91\% (N=10) elected to have schoolwork sessions with the SAR companion system under condition C when not required to do so. Participants spent an average of 93.041 minutes (SE = 19.065) in active sessions with the study companion robot under condition C. Figure \ref{fig:users_per_day} shows the number of users that started active sessions each day in the voluntary condition. Figure \ref{fig:sessions_per_hour} shows the distribution of session start times in each of the three conditions.

Results of the Wilcoxon signed-rank tests showed no significant increase between participants' pre-study and post-study ESQR scores (t = 0.81, p = 0.86), indicating that the minimal robot design had no measurable effect on the participant's executive functions. 

There was no significant difference between each participant's average post-session NASA-TLX scores during week A (no robot present) and during weeks B and C combined (robot present) (t = 0.31, p = 0.76). There was also no significant change in NARS scores pre-, mid-, and post-experiment (F = 0.50, p = 0.61).

To confirm that the order of the study conditions did not affect participant outcomes, we computed the difference between pre- and post-study scores on the NARS and ESQR for each participant and compared outcomes between the two ordering groups (group 1: cond. A $\rightarrow$ B $\rightarrow$ C, $n=6$) (group 2: cond. B $\rightarrow$ C $\rightarrow$ A, $n=5$). We performed Mann-Whitney U Tests and found no significant difference between groups 1 and 2  (ESQR: p = 0.5219; NARS: p = 0.2343), indicating that ordering effects were not present.

\subsection{Qualitative Results}

In this section, we report the themes that emerged from the post-study interviews and how they address RQ2 and RQ3. 

\subsubsection{Studying Behavior and Attitude Toward Studying}

Participants listed a wide variety of ways that Blossom affected their studying. Table \ref{tab:study_effects} shows the most common ideas that participants expressed. We broadly categorize the most common themes in this area as related to the participant's \textbf{ability to focus}, \textbf{motivation to study}, and \textbf{ability to manage their time}. 

\begin{table}
  \caption{Participant responses related to their behavior during study sessions and overall attitude toward studying}
  \label{tab:study_effects}
  \small
  \begin{tabular}{p{2.4in}p{0.6in}}
    \toprule
    \bf Statement& \bf Participants\\ \hline
    \midrule
    \bf Ability to focus\\ \hline
    Were distracted by the robot's jarring or startling movements& 6 (55\%) \\
    Found they were less distracted while studying& 5 (45\%) \\
    Noises made while the robot was moving and/or by the motors were distracting& 3 (27\%) \\
    Perceived Blossom as being disappointed when they got distracted; focused more on their work to avoid disappointing Blossom& 2 (18\%) \\
    The act of trying not to pay attention to the robot's movements made them more focused on their work & 2 (18\%)\\ \hline
    \bf Motivation to study\\ \hline
    Having Blossom's subtle companionship encouraged and motivated them to study & 3 (27\%) \\
    Found it easier to begin studying because Blossom's novelty made studying a fun and exciting activity& 3 (27\%) \\
    Interacting with the robot made studying more fun, thereby motivating them to study & 2 (18\%) \\\hline
    \bf Time Management \\ \hline
    Felt that they were better able to decide which order to complete tasks in and how to break tasks up into smaller chunks& 5 (45\%) \\
    Found they were more efficient and could get more done with Blossom&  4 (36\%) \\
    Found their work time was more structured with Blossom&  4 (36\%)  \\ 
    Felt better able to manage time because they were more focused and engaged& 3 (27\%) \\ 
    \bottomrule
  \end{tabular}
\end{table}

Participants suggested that a sense of companionship with the robot made it easier for them to study. For example, P4 compared studying with Blossom to being in a library: \textit{"it's kind of like when you're working at a library and you see everyone working around you, doing little movements, having their iPads out."} Concerning staying focused during study sessions, participants stated that Blossom's constant movement kept them engaged and that trying not to pay attention to the robot's movements made them pay more attention to their work. P5 stated \textit{"It was making some weird movements, but that has inspired me, it has motivated me to concentrate more on my studies and not get distracted by the movements. So that has, you know, that has improved my willpower that I should not concentrate on the robot."}
Participants also thought it was easier to focus when they could verbalize their thoughts to the robot. P7 stated \textit{"I talk a lot to myself while I do my work, so it was kind of fun to just talk, even though I know Blossom couldn't talk back. It made me feel like I was able to be more on task. I felt like my thoughts came more fluidly, and I felt more comfortable because I guess it wasn't a person there, but [the robot] made me feel like I wasn't completely talking to a wall. I just kind of felt like there's another presence, I guess."}
 
Blossom also positively impacted the participants' self-reported motivation to study. They expressed that interacting with Blossom made the schoolwork more enjoyable and that Blossom's subtle companionship and presence encouraged them to spend more time on schoolwork. When asked if it was easier to start working with the robot compared to the system with no robot, P3 explained that \textit{"I felt like it was slightly easier because it was more like a fun activity, turning on the system and having that robot next to me doing its little thing while studying. So I would say it was definitely more enjoyable [than the system with no robot] and I made sure that I was getting a few hours of work done every day."}

Finally, participants suggested that the study companion led them to manage their study time effectively. Blossom reminded them to use time wisely, which led them to take fewer unnecessary breaks. P1 stated \textit{"I started noticing I was like, 'I know I'm gonna get a lot of work done in the time that [Blossom]'s on.' So I'd be like, 'Okay, cool. I'm gonna sit down, I'm gonna get my work done, I'm gonna complete the session and my homework.' So I was starting to think, 'Oh, she's helping.'"} P7 reported \textit{"I don't think this was intended, but sometimes when I felt like I got off track, I would look over at it, and it seemed like it was looking at me being off track, not doing my work, so I was like, okay, I'll get back to doing my work."}

\subsubsection{General Feedback}

The most common positive feedback from participants related to Blossom's soft, animal-like appearance. The most common criticisms relate to Blossom's loud noises and jerky motions. \ref{tab:general_feedback} outlines the major themes from their responses.

\begin{table}
  \caption{Positive and negative feedback of the SAR study companion}
  \label{tab:general_feedback}
  \small
  \begin{tabular}{p{2.4in}p{0.6in}}
    \toprule
    \bf Statement& \bf Participants\\ \hline
    \midrule
    \bf Positive Features\\ \hline
    Found Blossom to be cute, friendly, or pet-like & 9 (82\%) \\
    Liked the zoomorphic design of Blossom and the crochet cover & 4 (36\%) \\
    Liked Blossom's small size & 2 (18\%) \\
    Liked Blossom's companionship, having someone to talk to & 2 (18\%) \\ \hline
    \bf Criticisms\\ \hline
    Disliked Blossom's loud motor noises&  8 (73\%) \\
    Found Blossom's movements distracting, jerky, random&  6 (55\%) \\
    Felt that Blossom and its setup took too much space or was not portable enough&  3 (27\%) \\
    \bottomrule
  \end{tabular}
\end{table}

Regarding positive feedback that did not directly relate to Blossom's performance during schoolwork sessions, participants frequently cited Blossom's cute, friendly, pet-like appearance, soft, crocheted cover, and small size and things they liked about the robot.
P1 stated, \textit{"I actually really enjoyed using it. I think it was just cute to have. It was like another presence."}
P4 stated, \textit{"I guess from a visual standpoint, it looks kind of like a pet, so I think that is a pretty good design choice. I think the ears are a good touch."}

The most common criticism of the SAR study companion related to Blossom's movements. Participants frequently reported disliking the jerky nature of Blossom's movements, which they found distracting. They also found the noise produced by the robot's servo motors distracting. Participants disliked that the system, including the robot, Raspberry Pi, and touch screen, was too large for some desks and too bulky to be relocated and utilized in other spaces. For example, participants said \textit{"the robot is easy to use except that it makes this motor noise when it makes the movements. I found that a bit distracting, but you get used to it if you use it for longer."} (P2) and \textit{"But yeah, he also takes up a lot of space, so I do have a fairly small desk... All those cables and stuff, it'd be nice to move them around the desk if I wanted to"} (P12). 

\subsubsection{Suggested Improvements and Ideas}

Participants gave a wide variety of suggestions for system improvements and new functionalities. Table \ref{tab:suggestions} outlines the major themes.

\begin{table}
  \caption{Suggestions and Ideas for the SAR Study Companion}
  \label{tab:suggestions}
  \small
  \begin{tabular}{p{2.4in}p{0.6in}}
    \toprule
    \bf Statement& \bf Participants\\ \hline
    \midrule
    \bf Suggestions\\ \hline
    Add assistant-like features, such as reminders, calendaring, assignment tracking, and general AI assistance & 9 (82\%) \\
    Enable Blossom to monitor attention and detect user distraction& 8 (73\%) \\
    Replace the touch screen that comes with Blossom with either physical buttons or a mobile app & 5 (45\%) \\
    Have Blossom provide affirmations during sessions & 2 (18\%) \\
    \bottomrule
  \end{tabular}
\end{table}

The most common suggestion related to enabling the robot to sense the user's emotional state or periods of distraction and respond. Participants suggested that Blossom could detect when they were distracted and use some cue to recapture their attention. 
P10 suggested \textit{"Maybe it could also incorporate a camera. So the other day I saw that you can capture emotions. Maybe for different emotions, you can give specific voice notes to the person. So like if they are feeling maybe distracted, the robot could say something that could motivate them or something like that."}

Conversely, some participants thought that intense monitoring would be unnecessary or unwelcome. P12 stated \textit{"When you study with a pet, you know, they can't talk, they're just there and Blossom's just there and I feel like for me that was enough."}

When asked to ideate new functions they would find helpful in an in-dorm robot, participants gave many suggestions to improve Blossom and its corresponding system. They suggested adding digital assistant-like features to Blossom, such as reminders, calendaring, assignment tracking, and general AI assistance. P4 stated, \textit{"I don't want to say like an AI, like where you ask a question and [the system] answers it, but something similar to that, kind of like Siri or Alexa where you can ask your questions, or maybe do some note-taking things, or even simple things that stop you from looking at your phone, like to check the time or check the weather."} P9 suggested interactions to help users consolidate and organize their thoughts as a potential function: \textit{"I could just like give it a bunch of ideas and be like, Okay, here's my thought process. Can you just help me out in that way? So, I guess kind of like when you talk to a TA, but like on demand would be really cool."}

Regarding the design and appearance of the robot, participants suggested that the touch screen be replaced with a mobile app or physical buttons. P1 suggested \textit{"I was more the little screen that I was trying to move. I think if it would be possible, maybe do the interface on a mobile app if that's somehow still connected to her, so it's Blossom herself and not the [touch screen] as well."} Others wanted Blossom to be made more portable so that it could be used in other rooms or common spaces. P12 suggested \textit{"I think I would really like to just have him as, maybe like a pet; like have him on my desk and then move him over to the kitchen when I'm cleaning, just to be there."}

\subsubsection{Influence of Recording and Study Procedure}

To isolate the effect of the robot from the effect of being recorded, we asked participants if their behavior was impacted by the presence of the webcam and the knowledge that they were being recorded. Three participants (27\%) disclosed that Blossom's video and audio recording capabilities made them more aware of their actions and more attentive toward work (P3, P4, P5). Eight participants (73\%) reported behaving differently due to being recorded: participants did not use their phones during active sessions (P1, P5, P10), felt added pressure to focus (P1, P2, P3, P4, P12), and suppressed their normal behaviors (such as talking aloud) because they felt self-conscious in front of the camera (P3, P4). However, out of the 10 participants who chose to complete sessions with the robot in condition C, only two opted to close the camera cover during these sessions.

\section{Discussion}

This work proposed a minimally interactive initial design of an in-dorm SAR study companion to support college students with ADHD in performing schoolwork tasks. The robot's perceived usability was evaluated through an in-dorm user study spanning multiple weeks with participants sampled from the intended user population, college students with ADHD. Relative to each research question, we found that: 1) participants demonstrated that the system was usable, even with minimal functionality; 2) participants found that the robot enacted a sense of companionship and accountability, but found the noise of the robot's motors and jerky movements distracting; and 3) participants gave a variety of suggestions to extend the functionality of the study companion robot. Next, we discuss the results relative to each of the research questions. 

\vspace{5pt}
\noindent{\it {\bf RQ1: Perceived Usefulness}}\newline
The average SUS score of 83.8 and the continued use of the robot when no longer required by all but one of the study participants indicate that our sample of college students with symptoms of ADHD found Blossom useful as a study companion, even with minimal, non-interactive behavior. The single participant who did not start a schoolwork session with Blossom under condition C reported in their post-study survey that they traveled during that week of the study. For this reason, we recommend that researchers hoping to employ a similar structured-to-unstructured study design take care to directly confirm with participants that they will be living in their dorm for the entire duration of the study, as participants may interpret the lack of requirements in the final condition as permission to travel or take other actions that otherwise prevent system use. From the distribution of session start times in Figure \ref{fig:sessions_per_hour}, we can see that participants rarely started sessions during typical workday hours and instead chose to initiate sessions in the evenings, as late as 1-3 am. One potential interpretation is that Blossom can fill a desire for companionship at times when another person may not be available.

\vspace{5pt}
\noindent{\it {\bf RQ2: Feedback on Within-Session Robot Performance}}\newline
When commenting on interactions with the SAR study companion, participants reported that they enjoyed Blossom's subtle movements, yet the jerky nature of the robot's movements and the loud noises that accompanied them were the most common complaints about the interaction. Because participants reported liking the movements, seemingly despite these negative qualities, it is likely that Blossom's idle motions were not inherently distracting to users with ADHD, but that implementing those motions with noisy actuators at high speed or too frequently could be more distracting than helpful. Even with the very loud motors in the present system, participants reported "getting used to" the noises quickly, and, in some cases, the added noise fortified them to stay focused on their work.

\vspace{5pt}
\noindent{\it {\bf RQ3: Feedback on Added Functionality}}\newline
Based on their feedback, participants appreciated Blossom's zoomorphic appearance, soft exterior, and small size. They generated a wide array of ideas for useful features that could be added to the robot that spanned from those highly related to study companions (e.g., reminders, assignment tracking) to those beyond the problem space of schoolwork (e.g., functionality as a cooking assistant (P6) or coffee-maker (P5). The repeated themes of removing the bulky touch screen and wanting the robot to be more portable inform our immediate next steps in the system design process. We will explore replacing the Raspberry Pi-based UI with a mobile phone app or incorporating a built-in battery pack as a power source. Given the widespread positive feedback regarding Blossom's appearance, we plan to continue to use Blossom in future iterations of the study companion design with the modifications described above.

\subsection{Limitations and Future Work}
In all conditions, factors beyond compensation may have motivated them to use the robot when they otherwise would not have done so. Participants may have continued completing daily sessions if they did not see the email reminder to begin condition C or if they intended to "make up" missed requirements for condition B in the previous week. Including a session counter and information about the study condition on the system's UI display would reduce some of the confounds about participant motivation.
Experimenter demand effects ~\cite{zizzo2010experimenter, nichols2008good} may have also played a role in the observed results; participants may have used the robot with the belief that the researchers wanted or expected them to do so. One way to limit this effect in future studies is to refrain from video- and audio-recording the sessions to minimize participants' beliefs that the researchers will know when they have not used the system.

Although our study provided preliminary insights about the viability of study companion robots, future work should investigate how the robot's behavior affects the study session interaction. Future research should explore a condition in which the robot is present but does not display any movement to determine whether the voluntary use observed in our study can be attributed to the robot's idle motions. 
Finally, future work will include the analysis of the study session recordings. In line with participants' suggestions to equip the robot with user monitoring capabilities, we intend to analyze visual and audio features to understand how participants' behavior differed between conditions and how multi-modal data can be used to predict the user's state during a study session.

\section{Conclusion}
This work contributes the participatory design and evaluation of a socially assistive robot study companion for college students with ADHD. Our findings show that: RQ1) A sample of college students with ADHD found the system useful, even in its initial pilot state, and elected to continue using it when they were no longer required to; RQ2) College students with ADHD have demonstrated use for and interest in in-dorm SAR study companions, although users have differing and unique preferences for the robot's behavior in this role; In addition, users found the loud noises and abrupt movements of the robots' servomotors distracting during a study session, but they were able to acclimate with repeated use; and RQ3) test users proposed many ideas for potential improvements to the robot's design and additional features that they would find useful in an in-dorm robot. These findings suggest that in-dorm robots have potential as long-term assistive devices for college students with ADHD. Furthermore, we demonstrated the feasibility of incorporating a long-term in-dorm deployment into the early stages of a participatory design process for Human-Robot Interactions.

\begin{acks}
This research was supported by the National Science Foundation Grant NSF IIS-1925083 and the NSF REU program.
The authors extend additional thanks to Caroline Kenney and Anna-Maria Valentza. 
\end{acks}

\bibliographystyle{ACM-Reference-Format}
\balance
\bibliography{base}


\end{document}